\pdfoutput=1
\documentclass[11pt]{article}

\usepackage[]{acl}

\usepackage{times}
\usepackage{latexsym}
\usepackage{amsmath,amssymb}
\usepackage{mathtools}
\usepackage{multirow}
\usepackage[ruled,linesnumbered]{algorithm2e}

\usepackage[T1]{fontenc}
\usepackage[utf8]{inputenc}

\usepackage{microtype}

\usepackage{amsmath}
\usepackage{times}
\usepackage{latexsym}
\usepackage{colortbl}
\usepackage{multirow}
\usepackage[pdftex]{graphicx}
\usepackage{caption}
\usepackage{mathtools}
\usepackage{subfigure}
\usepackage{amsfonts}
\usepackage{physics}

\title{FAMIE: A Fast Active Learning Framework for Multilingual Information Extraction}

\author{Minh Van Nguyen\textsuperscript{\rm 1}, Nghia Trung Ngo\textsuperscript{\rm 1},
Bonan Min\textsuperscript{\rm 2}, and Thien Huu Nguyen\textsuperscript{\rm 1} \\
\textsuperscript{\rm 1} Dept. of Computer and Information Science, University of Oregon, Eugene, OR, USA\\
\textsuperscript{\rm 2} Raytheon BBN Technologies, USA \\
  \texttt{\{minhnv@cs,nghian@,thien@cs\}.uoregon.edu}, \\ \texttt{bonan.min@raytheon.com}
}

\begin{document}
\maketitle

% =================================
% ABSTRACT
% =================================
\begin{abstract}
This paper presents FAMIE, a comprehensive and efficient active learning (AL) toolkit for multilingual information extraction. FAMIE is designed to address a fundamental problem in existing AL frameworks where annotators need to wait for a long time between annotation batches due to the time-consuming nature of model training and data selection at each AL iteration. This hinders the engagement, productivity, and efficiency of annotators. Based on the idea of using a small proxy network for fast data selection, we introduce a novel knowledge distillation mechanism to synchronize the proxy network with the main large model (i.e., BERT-based) to ensure the appropriateness of the selected annotation examples for the main model. Our AL framework can support multiple languages. The experiments demonstrate the advantages of FAMIE in terms of competitive performance and time efficiency for sequence labeling with AL. We publicly release our code (\url{https://github.com/nlp-uoregon/famie}) and demo website (\url{http://nlp.uoregon.edu:9000/}). A demo video for FAMIE is provided at: \url{https://youtu.be/I2i8n_jAyrY}.

\end{abstract}

% =================================
% INTRODUCTION
% =================================
\section{Introduction}

Information Extraction (IE) systems provide important tools to extract structured information from text \cite{Li:14,Nguyen:19,lai2021graph,veyseh2021augmenting,nguyen-etal-2021-cross}. At the core of IE involves sequence labeling tasks that aim to recognize word spans and semantic types for some objects of interest (e.g., entities and events) in text. For example, two typical sequence labeling tasks in IE feature Named Entity Recognition (NER) to find names of entities of interest, and Event Detection (ED) to identify triggers of specified event types \cite{Walker:05}. Despite extensive research effort for sequence labeling \cite{lafferty2001conditional,Ma:16,pouran-ben-veyseh-etal-2021-modeling}, a major bottleneck of existing IE methods involves the requirement for large-scale human-annotated data to build high-quality models. As annotating data is often expensive and time-consuming, large-scale labeled data is not practical for various domains and languages.

%Due to their importance, extensive research effort has been devoted to develop effective models for sequence labeling tasks in IE \cite{lafferty2001conditional,Ma:16,devlin2018bert}. However, the major bottleneck of existing IE methods involves the requirement for large-scale human-annotated data that is necessary to build high-quality models for applications. As human-annotated data is often very expensive and time-consuming to achieve for sequence labeling tasks in IE, the large-scale labeled data requirement is not practical for various domains and languages.

%, thus hindering the applicability of current sequence labeling methods for IE in various domains and languages.

%current sequence labeling methods for IE have not been widely applied to build real-world applications for different domains and languages. T

%scarcity or unavailability of human-annotated data in various languages/domains, thus hindering the development of high-quality IE systems for applications.

%a small set of annotated examples and

To address the annotation cost for IE, previous work has resorted to active learning (AL) approaches \cite{settles-craven-2008-analysis,settles2009active} where only a selective set of examples are annotated to minimize the annotation effort while maximizing the performance. Starting with a set of unlabeled data, AL methods train and improve a sequence labeling model via multiple human-model collaboration iterations. At each iteration, three major steps are performed in order: (i) training the model on the current labeled data, (ii) using the trained model to select the most informative examples in the current unlabeled set for annotation, and (iii) presenting the selected examples to human annotators to obtain labels. In AL, the number of annotated samples or annotation time might be limited by a budget to make it realistic.

%Accordingly, the selection algorithm in the second step should select the most informative examples to optimize performance given the annotation budget. 

Unfortunately, despite much potentials, existing AL methods and frameworks are still not applied widely in practice due to their main focus on devising the most effective example selection algorithm for human annotation, e.g., based on the diversity of the examples \cite{shen2017deep,yuan-etal-2020-cold} and/or the uncertainty of the models \cite{roth2006margin,wang2014new,shelmanov-etal-2021-active}. Training and selection time in the first and second steps of each AL interaction is thus not considered in prior work for sequence labeling. This is a critical issue that limits  the application of AL: annotators might need to wait for a long period between annotation batches due to the long training and selection time of the models at each AL iteration. Given the widespread trend of using large-scale pre-trained language models (e.g., BERT), this problem of long waiting or training/selection time in AL can only become worse. On the one hand, the long idle time of annotators reduces the number of annotated examples given an annotation budget. Further, the engagement of annotators in the annotation process can drop significantly due to the long interruptions between annotation rounds, potentially affecting the quality of their produced annotation. In all, current AL frameworks are unable to optimize the available time of annotators to maximize the annotation quantity and quality for satisfactory performance.

%given an annotation budget.

To this end, we demonstrate a novel AL framework (called FAMIE) that leverages large-scale pre-trained language models for sequence labeling to achieve optimal modeling capacity while significantly reducing the waiting time between annotation rounds to optimize annotator time. Instead of training the full/main large-scale model for data selection at each AL iteration, our key idea is to train only a small proxy model on the current labeled data to recommend new examples for annotation in the next round. In this way, the training and data selection time can be reduced significantly to enhance annotation engagement and quality. An important issue in this idea is to ensure that the examples selected by the proxy model are also optimal for the main large model. To this end, we introduce a novel knowledge distillation mechanism for AL that encourages the synchronization between the proxy and main models, and promotes the fitness of selected examples for the main model. To update the main model with new annotated data for effective distillation, we propose to train the main large model on current labeled data during the annotation time, thus not adding to the waiting time of annotators between annotation rounds. This is in contrast to previous AL frameworks that leave the computing resources unused during annotation time. Our approach can thus efficiently exploit both human and computer time for AL.

%To ensure the effectiveness of the suggested examples for the final full model, the small proxy model will also be synchronized with the full large model via knowledge distillation. In this way, the training and data selection time can be reduced significantly while still maintaining the advantage of selection algorithms for AL. Crucially, to 
    
%In addition, to further speed up the training of the two models, we propose to use light-weight adapter networks that are inserted into large pre-trained models. During the training, we only update the adapter networks while the parameters of the original pre-trained models are fixed. Adapter networks have been shown to achieve similar performance for NLP tasks compared to the standard update of all model parameters \cite{houlsby2019parameter,peters2019tune,pfeiffer2020adapterhub}. 

To evaluate the proposed AL framework FAMIE, we conduct experiments for multilingual sequence labeling problems, covering two important IE tasks (i.e., NER and ED) in three languages (i.e., English, Spanish, and Chinese). The experiments demonstrate the efficiency and effectiveness of FAMIE that can achieve strong performance with significantly less human-computer collaboration time. Compared to existing AL systems such as  ActiveAnno \cite{wiechmann-etal-2021-activeanno} and Paladin \cite{nghiem-etal-2021-paladin}, our system FAMIE features important advantages. First, FAMIE introduces a novel approach to reduce model training and data selection time for AL via a small proxy model and knowledge distillation while still benefiting from the advances in large-scale language models. Second, while previous AL systems only focus on some specific task in English, FAMIE can support different sequence labeling tasks in multiple languages due to the integration of our prior multilingual toolkit Trankit \cite{nguyen2021trankit} to perform fundamental NLP tasks in 56 languages. Third, in contrast to previous AL systems that only implement one data selection algorithm, FAMIE covers a diverse set of AL algorithms. Finally, FAMIE is the first complete AL system that allows users to define their sequence labeling problems, work with the models to annotate data, and eventually obtain a ready-to-use model for deployment.

\section{System Description}

% \subsection{Active Learning for Sequence Labeling}
In AL, we are given two initial datasets, a small seed set of labeled examples $D_0=\{(\textbf{w}, \textbf{y})\}$ and an unlabeled example set $U_0=\{\textbf{w}\}$ (the seed set $D_0$ is optional and our system can work directly with only $U_0$).
For sequence labeling, models consume a sequence of $K$ words $\textbf{w}=[w_1, w_2, \ldots, w_K]$ (i.e., a sentence/example) to output a tag sequence $\textbf{y}=[y_1, y_2, \ldots, y_K]$ ($y_i$ is the label tag for $w_i$). The tag sequence is represented in the BIO scheme to capture spans and types of objects of interest. 

A typical AL process contains multiple rounds/iterations of model training, data selection, and human annotation in a sequential manner. Let $D$ and $U$ be the overall labeled and unlabeled set of examples at the beginning of the current $t$-th iteration (initialized with $D_0$ and $U_0$). At the current iteration, a sequence labeling model is first trained on the current labeled set $D$. A sample selection algorithm then employs the trained model to suggest the most informative subset of examples $U^t$ in $U$ (i.e., $U^t \subset U$) for annotation. Afterwards, a human annotator will provide labels for the sentences in the selected set $U^t$, leading to the labeled examples $D^t$ for $U^t$. The labeled and unlabeled sets can then be updated via: $D \leftarrow D \cup D^t$ and $U \leftarrow U \setminus U^t$.

\begin{figure}
\addtolength{\belowcaptionskip}{-3mm}
  \centering
% \captionsetup{skip=5pt}
  \includegraphics[width=0.4\textwidth]{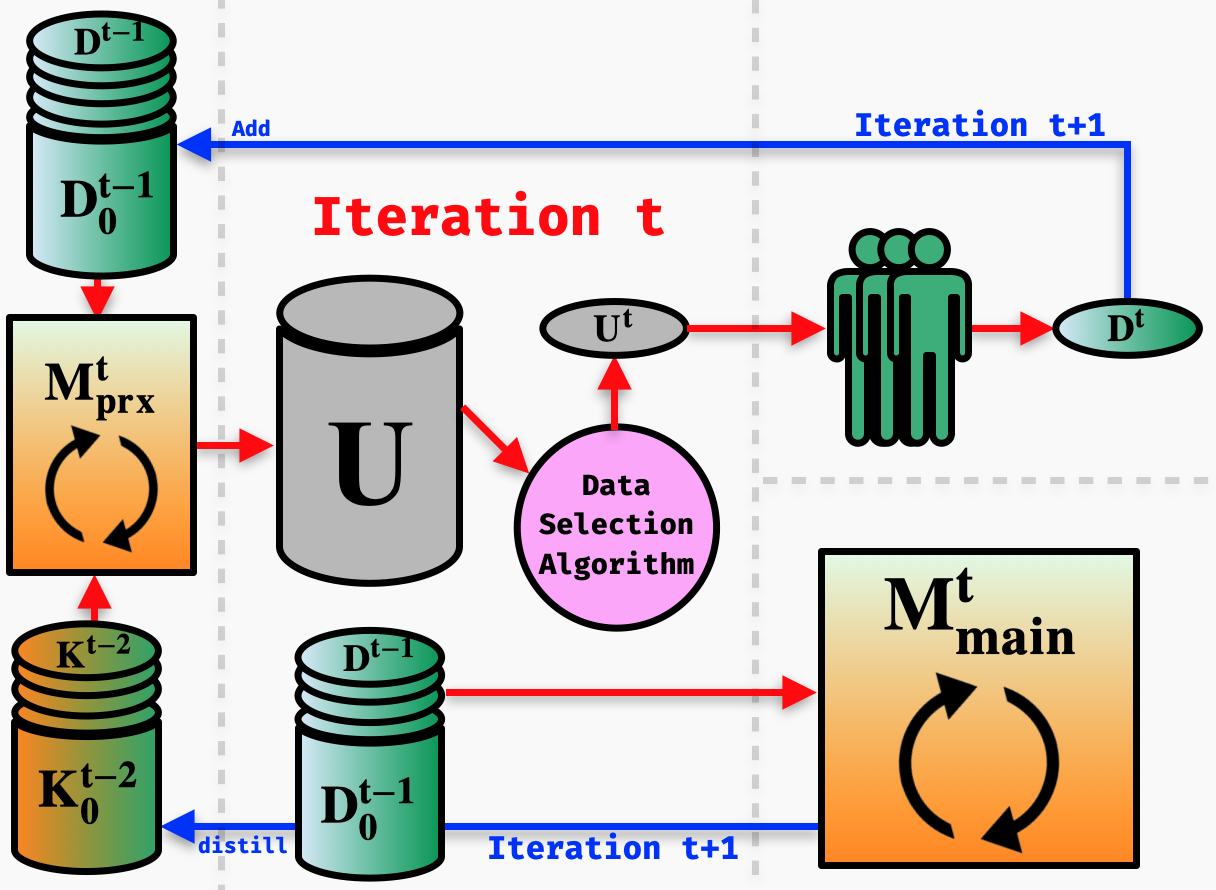}
  \caption{\small The overall Proxy Active Learning process.}
%   \caption{\small The overall Proxy Active Learning process. \textbf{(Left)} unlabeled data is queried by selection algorithm from pool based on trained proxy model. \textbf{(Middle)} annotators label select instances while target model is being retrained simultaneously. \textbf{(Right)} newly acquired labels are combined with the pre-computed distillation signals to retrain to proxy model.} 
\label{fig:1}
\end{figure}
    
\subsection{Model}
We employ the typical Transformer-CRF architecture for sequence labeling \cite{nguyen2021trankit}. In particular, given the input sentence $\textbf{w}=[w_1, w_2, \ldots, w_K]$, the state-of-the-art multilingual language model \textrm{XLM-Roberta} \cite{conneau-etal-2020-unsupervised} is used to obtain contextualized embeddings for the words: $\textbf{X} = \textbf{x}_1, \ldots, \textbf{x}_K = \textrm{XLMR}(w_1, \ldots, w_K)$ (i.e., to support multiple languages). Afterwards, the word embeddings are sent to a feed-forward network with softmax in the end to obtain the score vectors: $\textbf{z}_i = \textrm{softmax}(\textbf{h}_i)$ where $\textbf{h}_i = \textrm{FFN}(\textbf{x}_i)$. Here, each value in $\textbf{z}_i$ represents a score for a tag in the tag set $V$. The score vectors are then fed into a Conditional Random Field (CRF) layer to compute a distribution for possible tag sequences for $\textbf{w}$: $P(\hat{\textbf{y}}|\textbf{w}) = \frac{\textrm{exp}(s(\hat{\textbf{y}}, \textbf{w}))}{\sum_{\hat{\textbf{y}}' \in Y(\textbf{w})}\textrm{exp}(s(\hat{\textbf{y}}', \textbf{w}))}$ where $Y(\textbf{w})$ is the set of all possible tag sequences for $\textbf{w}$. Also, $s(\hat{\textbf{y}}, \textbf{w})$ is the score for a tag sequence $\hat{\textbf{y}} = [\hat{y}_1, \ldots, \hat{y}_K]$: $s(\hat{\textbf{y}}, \textbf{w}) = \sum_i \textbf{z}_i[\hat{y}_i] + \sum_i \pi_{\hat{y}_{i}\rightarrow \hat{y}_{i+1}}$. Here, $\pi_{\hat{y}_{i}\rightarrow \hat{y}_{i+1}}$ is the transition score from the tag $\hat{y}_{i}$ to the tag $\hat{y}_{i+1}$. The model is trained by minimizing the negative log likelihood: $L_{task} = - \log P(\textbf{y}|\textbf{w})$.
For inference, the Viterbi algorithm is used for decoding: $\hat{\textbf{y}^*} = \textrm{max}_{\hat{\textbf{y}}'} P(\hat{\textbf{y}}'|\textbf{w})$.

% s(\textbf{y}, \textbf{w}) + \textrm{log}\sum_{\hat{\textbf{y}}' \in Y(\textbf{w})}\textrm{exp}(s(\hat{\textbf{y}}', \textbf{w}))

\paragraph{Adapter-based Finetuning} To further improve the memory and time efficiency, we incorporate light-weight adapter networks \cite{houlsby2019parameter,peters2019tune} into our model. In form of small feed-forward networks, adapters are injected in between the transformer layers of \textrm{XLM-Roberta}. For training, we only update the adapters while the parameters of \textrm{XLM-Roberta} are fixed. This significantly reduces the amount of learning parameters while sacrificing minimal extraction loss, or in case of low-resource learning even surpassing performance of fully fine-tuned models.
    
\subsection{Data Selection Strategies}

%In contrast to previous AL framework that only implements one data selection approach \cite{wiechmann-etal-2021-activeanno,nghiem-etal-2021-paladin},

To improve the flexibility to accommodate different problems, our AL framework supports a wide range of data selection strategies for choosing the best batch of examples to label at each iteration for sequence labeling. These algorithms are categorized into three groups, i.e., uncertainty-based, diversity-based, and hybrid. For each group, we explore its most popular methods as follows.

%For all selection methods, we use the same Transformer-CRF model with {\it XLM-R} as the main sequence labeling architecture.

%, that take into account distinct criteria when comparing unlabeled examples.

%and might introduce different performance-efficiency trade-offs

% for the examples

\paragraph{Uncertainty-based.} These methods select examples for annotation according to the main model's confidence over the predicted tag sequences for unlabeled examples. Early methods sort the unlabeled examples by the uncertainty of the main model. To avoid the preference over longer examples, the method Maximum Normalized Log-Probability (\textbf{MNLP}) \cite{shen2017deep} proposes to normalize the likelihood over example lengths. In particular, MNLP selects examples with the highest MNLP scores: $MNLP(\textbf{w}) = - \textrm{max}_{\hat{\textbf{y}}'} \frac{1}{K} \textrm{log} P(\hat{\textbf{y}}' | \textbf{w})$.

%However, these methods usually favor longer examples.

%as they tend to have lower probabilities

\paragraph{Diversity-based.} Algorithms in this category assume that a representative set of examples can act as a good surrogate for the whole dataset.
\textbf{BERT-KM} \cite{yuan-etal-2020-cold} uses $K$-Means to cluster the examples in unlabeled data based on the contextualized embeddings of the sentences (i.e., the representations for the [CLS] tokens in the trained BERT-based models). The nearest neighbors to the $K$ cluster centers are then chosen for labeling. 

\paragraph{Hybrid.} Recently, several works have proposed data selection strategies for BERT-based AL to balance between uncertainty and diversity.
The \textbf{BADGE} method \cite{ash2019deep,kim-2020-deep} chooses examples from clusters of gradient embeddings, which are formed with the token representations $\textbf{h}_i$ from the penultimate layer of the main model and the gradients of the cross-entropy loss with respect to such token representations. The gradient embeddings are then sent to the $K$-Means++ to find the initial $K$ cluster centers that are distant from each other, serving as the selected examples \cite{kim-2020-deep}.

%In particular, a gradient embedding for an input sentence $\textbf{w}$ consists of $|V|$ sub-embeddings $\textbf{g}_{\textbf{w}} = [(\textbf{g}_{\textbf{w}})_1, ..., (\textbf{g}_{\textbf{w}})_{|V|}]$; each corresponds to a tag $t \in V$: $(\textbf{g}_{\textbf{w}})_t = \sum_i (p_i[t] - \hat{y}_i[t]) \textbf{h}_i$. Here, $p_i = P(y_i|\textbf{w}) := \textbf{z}_i$ is the individual label distribution for $w_i$ and $\hat{y}_i$ is the one-hot vector for the predicted tag for $w_i$.

%using the individual label distribution

%$\textbf{e}_{\textbf{w}}$

In addition, we implement the AL framework \textbf{ALPS} \cite{yuan-etal-2020-cold} that does not require training the main model for data section. ALPS employs the surprisal embedding of $\textbf{w}$, which is obtained from the likelihoods of masked tokens from pre-trained language models (i.e., \textrm{XLM-Roberta}). The surprisal embeddings are also clustered to select annotation examples as in BERT-KM.

\subsection{Proxy Active Learning}

As discussed in the introduction, model training and data selection at each iteration of traditional AL methods might consume significant time (especially with the current trend of large-scale language models), thus introducing a long idle time for annotators that might reduce annotation quality and quantity. To this end, \cite{shelmanov-etal-2021-active} have explored approaches to accelerate training and data selection steps for AL by leveraging smaller and approximate models during the AL iterations. To make it more efficient, the main large model is only trained once in the end over all the annotated examples in AL. Unfortunately, this approach suffers from the mismatch between the approximate and main models as they are separately trained in AL, thus limiting the effectiveness of the selected examples for the main model \cite{lowell-etal-2019-practical}. 

%. Even worse, the exclusion of the main model in the interaction with annotators for AL might be harmful for the performance of the main target model

To overcome these issues, our AL framework FAMIE trains a small proxy network at each iteration to suggest new unlabeled samples. Dealing with the mismatch between the proxy-selected examples and the main model, FAMIE proposes to involve the main model in the training and data selection for the proxy model. In particular, at each AL iteration, the main model will still be trained over the latest labeled data. However, to avoid the interference of the main large model with the waiting time of annotators, we propose to train the main model during the annotation time of the annotators (i.e., main model training and data annotation are done in parallel). Given the main model trained at previous iteration, knowledge distillation will be employed to synchronize the knowledge between the main and proxy models at the current iteration.

%, thus promoting the effectiveness of the selected examples.

%for the main model. 

%which parallelizes the labeling of annotators with the training of main model on the lastest labeled set.

%This is achieved by employing an additional small proxy model for selecting unlabeled data that will be 

The complete framework for FAMIE is presented in Figure \ref{fig:1}. At iteration $t$, a proxy acquisition model is trained on the current labeled data set $D_0^{t-1} = D^0 \cup D^1 \ldots \cup D^{t-1}$. The trained proxy model at the current step is called $M^t_{prx}$. Also, we use knowledge distillation signals $K^{t-2}_{0}$ that is computed from the main model $M^{t-1}_{main}$ trained at the previous iteration $t-1$ to synchronize the proxy model $M^t_{prx}$ and the main model $M^{t-1}_{main}$ ($M^1_{prx}$ is trained only on $D^0$). Afterwards, a data selection algorithm is used to select a batch of examples $U^t$ from the current unlabeled set $U$ for annotation, leveraging the feedback from $M^t_{prx}$. Next, a human annotator will label $U^t$ to produce the labeled data batch $D^t$ for the next iteration $t+1$. During this annotation time, the main model will also be trained again over the current labeled data $D_0^{t-1}$ to produce the current version $M^t_{main}$ of the model. The distillation signal $K^{t-1}_{0}$ for the next step will also be computed after the training of $M^t_{main}$. This process is repeated over multiple iterations and the last version of $M_{main}$ will be returned for users.

To improve the fitness of the proxy-based selected examples for $M_{main}$, we leverage the distilled version \textrm{miniLM} of \textrm{XLM-Roberta} \cite{wang-etal-2021-minilmv2} that employs similar stacks of transformer layers for the proxy model $M_{prx}$. Note that $M_{prx}$ also includes a CRF layer on top of \textrm{miniLM}.

\subsection{Uncertainty Distillation}
%While the fine-tuned proxy model $M^t_{prx}$ over $D$ might be able to extract important knowledge from the current labeled set, there might be still large mismatch between $M^t_{prox}$ and the main model's uncertainty on unlabeled set. In particular, the two models’ decision boundaries for a sequence labeling task can be different significantly even though they are trained on similar data. This prompts a demand for regularizing the proxy model's predictions to be consistent with those of a trained main model. Ideally, we expect the tag sequence distribution $P_{prx}(\textbf{y} | \textbf{w})$ learned by the proxy model to mimic the tag sequence distribution $P_{main}(\textbf{y} | \textbf{w})$ learned by the main model. This can be done via minimizing the KL divergence between the two distributions: $KL = - \sum_{\textbf{y}' \in Y(\textbf{w})} P_{main}(\textbf{y}' | \textbf{w}) \textrm{log}  P_{prx}(\textbf{y} | \textbf{w})$.

Although the proxy and main model $M_{prx}$ and $M_{main}$ are trained on similar data, they might still exhibit large mismatch, e.g., regarding decision boundaries. This prompts a demand for regularizing the proxy model's predictions to be consistent with those of a trained main model to improve the fitness of the selected examples for $M_{main}$. Ideally, we expect the tag sequence distribution $P_{prx}(\textbf{y} | \textbf{w})$ learned by the proxy model to mimic the tag sequence distribution $P_{main}(\textbf{y} | \textbf{w})$ learned by the main model. To this end, we propose to minimize the difference between the intermediate outcomes (i.e., the unary and transition scores) of the two distributions. In particular, we introduce the following distillation objective for each sentence $\textbf{w}$ at one AL iteration: $L_{dist} = -\sum_i \sum_v p^{main}_i[v] \log p^{prx}_i[v] + \sum_i (\pi^{main}_{y_{i}\rightarrow y_{i+1}} - \pi^{prx}_{y_{i}\rightarrow y_{i+1}})^2$ where $p^{main}_i$ and $p^{prx}_i$ are the tag distributions computed by the main and proxy models respectively for the word $w_i \in \textbf{w}$ (i.e., the scores $\textbf{z}_i$). Note that $p^{main}_i$ and $\pi^{main}_{y_{i}\rightarrow y_{i+1}}$ serve as the knowledge distillation signal that is obtained once the main model finishes its training at each iteration. Here, we will use the newly selected examples for the current annotation to compute the distillation signals. The overall objective to train $M_{prx}$ at each AL iteration is thus: $L = L_{task} + L_{dist}$.

\begin{figure}[ht!]
%\captionsetup{skip=2pt}
\addtolength{\belowcaptionskip}{-6mm}
\begin{center} 
\includegraphics[scale=0.36]{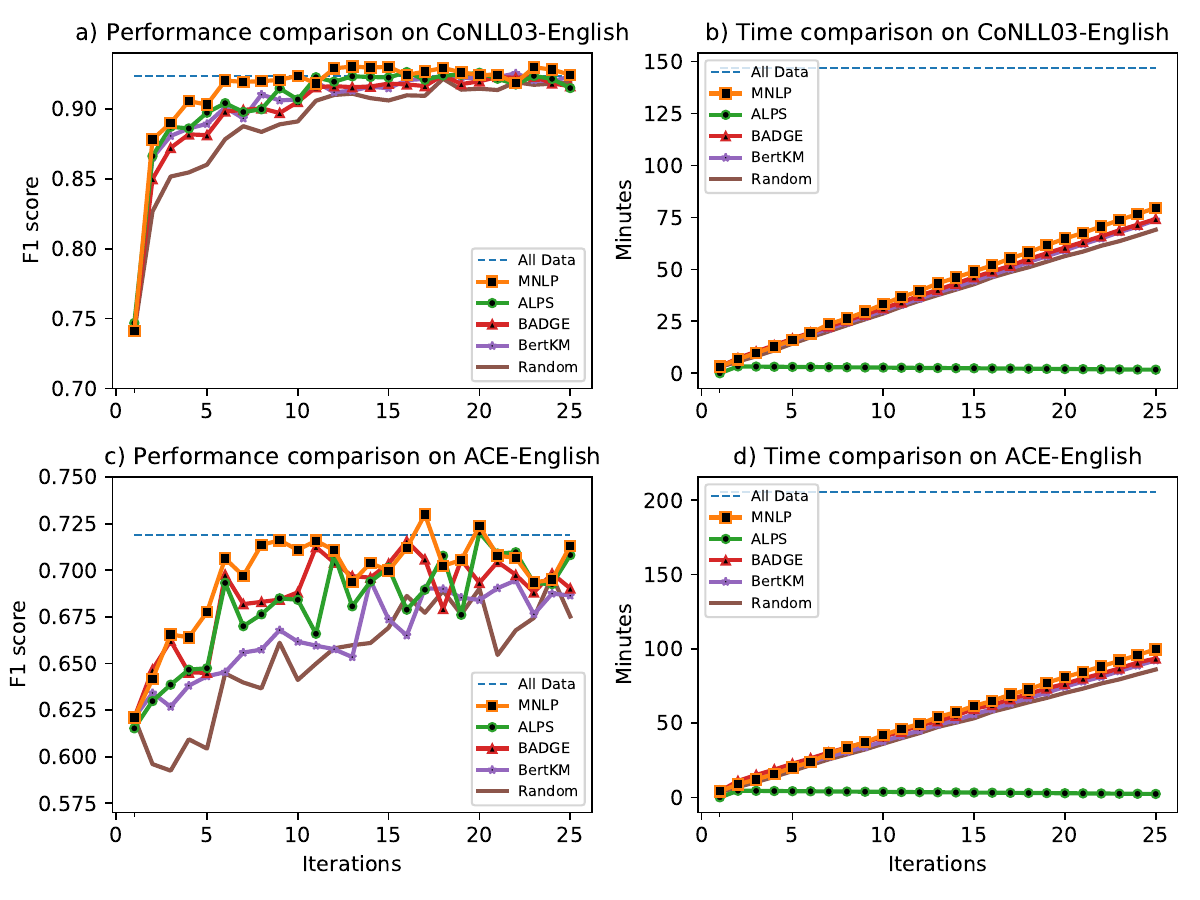}
\caption{\small Comparison among data selection strategies.} \label{fig:2}
\end{center}
\end{figure}

% \begin{figure}[ht!]
% \captionsetup{skip=2pt}
% \addtolength{\belowcaptionskip}{-6mm}
% \begin{center} 
% \includegraphics[scale=0.35]{figures/table2-english.pdf}
% \caption{\small Performance and time comparison between our proposed framework and the baselines.} \label{fig:3}
% \end{center}
% \end{figure}

% \begin{figure}[ht!]
% \captionsetup{skip=2pt}
% \addtolength{\belowcaptionskip}{-6mm}
% \begin{center} 
% \includegraphics[scale=0.35]{figures/table2-spanish-chinese.pdf}
% \caption{\small Performance and time comparison between our proposed framework and the baselines.} \label{fig:4}
% \end{center}
% \end{figure}

% =================================
% USAGE
% =================================
\section{Usage}
Detailed documentation for FaMIE is provided at: \url{https://famie.readthedocs.io/}. The codebase is written in Python and Javascript, which can be easily installed through PyPI at : \url{https://pypi.org/project/famie/}.

%is fully open-source and

%Once installed, users can start interactive AL processes to build models for their defined sequence labeling problems.

\noindent \textbf{Initialization.} To initialize a project, users first choose a data selection strategy and upload a label set to define a sequence labeling problem. Next, the dataset $U$ with unlabeled sentences should be submitted. FAMIE then allows users to interact with the models and annotate data over multiple rounds with a web interface. Also, FAMIE can detect languages automatically for further processing.

%users do not need to designate the specific language of their data as it can handle multilingual tasks automatically.

%annotation session on FaMIE web interface.
%Initially, annotators will be prompted to setup their session by choosing a task formulation (we currently support sequence labeling tasks) and one data selection strategy.
%A new project is then created that additionally requires users to upload their label sets and unlabeled datasets $U$ and labels to finalize the process.
%Note that there is no need for users to designate the specific language of their data as FaMIE is designed to handle multilingual tasks automatically.

%At the beginning of each iteration, FAMIE can minimize the waiting time of annotators for the next annotation batch as the proxy acquisition model $M_{prx}$ can be quickly trained and used to find the most beneficial examples from the current unlabeled pool $U$. 

\begin{figure}[ht!]
%\captionsetup{skip=5pt}
 \addtolength{\belowcaptionskip}{-8mm}
\begin{center} 
\includegraphics[scale=0.35]{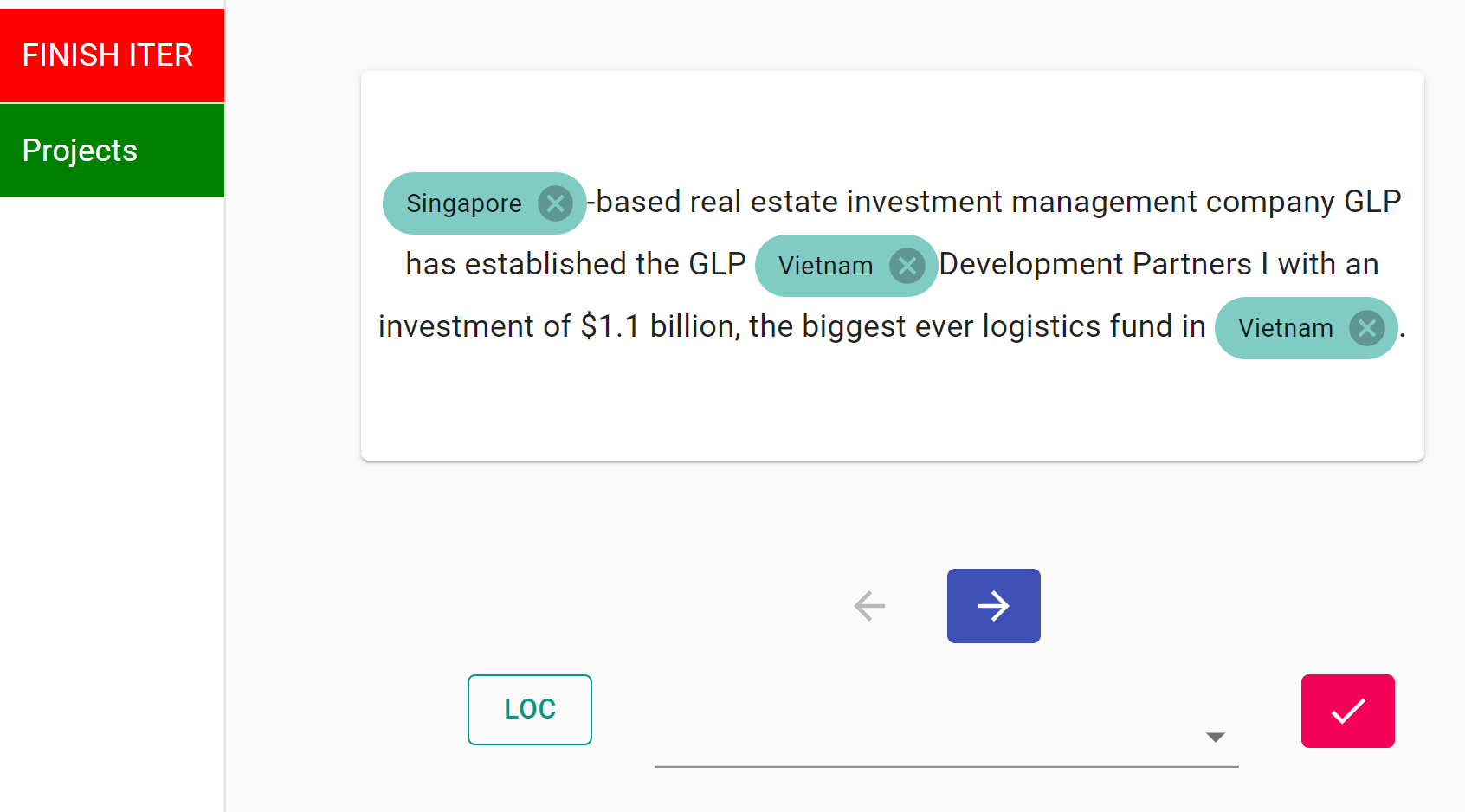}
\caption{\small Annotation interface in FAMIE.} \label{fig:ann}
\end{center}
\end{figure}

\noindent \textbf{Annotating procedure.} Given one annotation batch in an iteration, annotators label one sentence at a time as illustrated in Figure \ref{fig:ann}. In particular, the annotators annotate the word spans for each label by first choosing the label and then highlighting the appropriate spans. Also, FAMIE designs the size of the annotation batches to allow enough time to finish the training of the main model during the annotation time at each iteration.

%Simultaneously, the main model will be retrained on previous labeled set in the background.

%will finish learning as users complete their annotation iteration.

%At the beginning of each iteration, FAMIE will minimize the amount of idle time of users waiting for labeling candidates as the proxy acquisition model is quickly retrained and used to find the most beneficial examples from the current unlabeled pool.
%The selected instances are then pushed on the queue for annotators to query and start the next labeling process.
%The annotation is done sentence by sentence as illustrated in Figure. \ref{fig:5}. After choosing an label tag, users can annotate by highlighting appropriate spans for the label and continue to the next sentence when they think there is no more tagging needed for the current ones. 
%Simultaneously, the main model is being re-trained on previous labeled set in the background and will finish learning as users complete their annotation iteration.

\begin{figure}[ht!]
%\captionsetup{skip=5pt}
 \addtolength{\belowcaptionskip}{-7mm}
\begin{center} 
\includegraphics[scale=0.3]{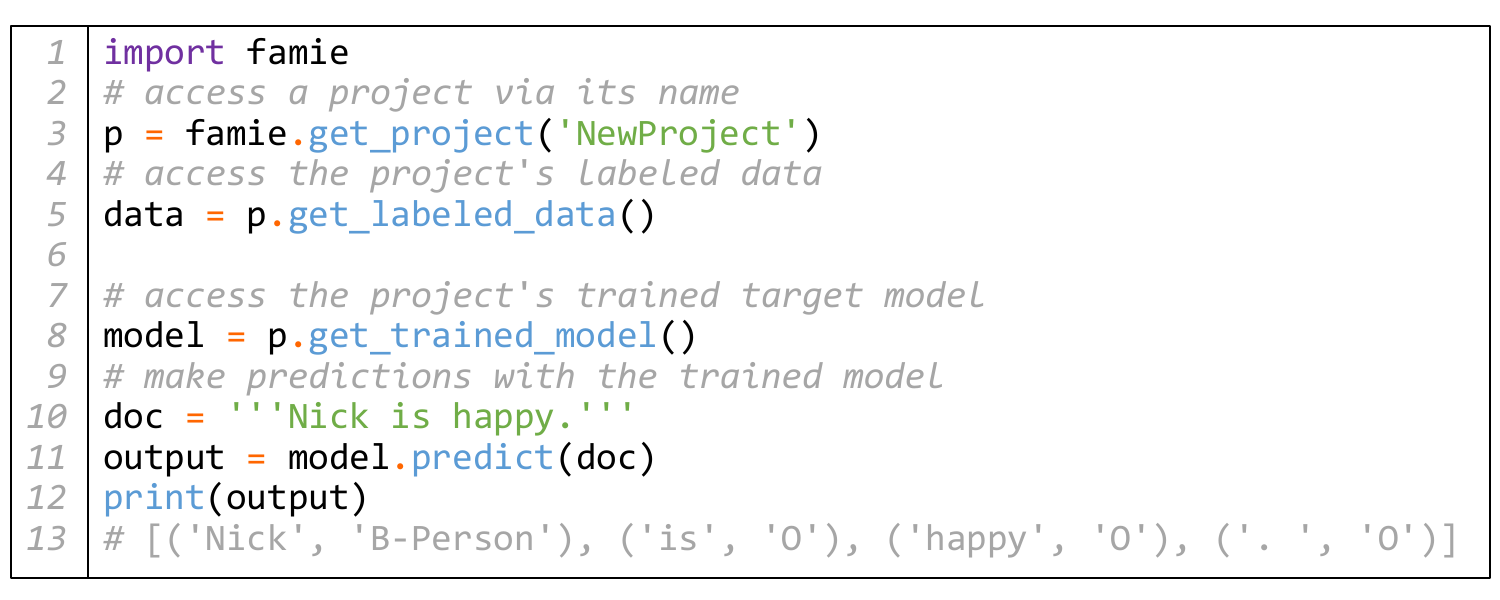}
\caption{\small Accessing the labeled dataset and the trained main model returned by an AL project.} \label{fig:code}
\end{center}
\end{figure}

\noindent \textbf{Output.} Unlike previous AL toolkits which focus only on their web interfaces to produce labeled data, FAMIE provides a simple and intuitive code interface for interacting with the resulting labeled dataset and trained main models after the AL processes. The code snippet in Figure \ref{fig:code} presents a minimal usage of our \textbf{famie} Python package to use the trained main model for inference over new data. This allows users to immediately evaluate their models and annotation efforts on data of interest.

% \begin{figure}[ht]
%     \centering
%     \frame{\includegraphics[scale=0.35]{figures/annotation-page-2.png}}
%     \caption{Annotation page.}
%     \label{fig:5}
% \end{figure}

% The labeled spans are then stored to the database when users click on the red button at the bottom right. The annotation continues for the next sentence by clicking on the blue arrow button. Once the annotation for the current iteration is complete, users can click on the \textit{``Finish Iter''} button at the top left corner to begin the next iteration.

% \noindent \textbf{Accessing labeled dataset and evaluating trained models} By default, the labeled dataset and the trained target model are stored locally at the users' machines after each annotation iteration. The labeled dataset and the trained target model can be accessed as below:

% \begin{figure}[ht]
%     \centering
%     \includegraphics[scale=0.3]{figures/basic-functions.pdf}
%     \caption{Accessing the labeled dataset and the trained target model.}
%     \label{fig:6}
% \end{figure}
% The labeled dataset is a list of labeled sentences, each sentence is a list of tokens along with their tagged labels. The trained model takes raw strings as inputs and outputs a list of tokens with their labels as shown in the figure.

\begin{table*}[htbp]
\center
%\small
%\captionsetup{skip=5pt}
% %\addtolength{\abovecaptionskip}{-3.0mm}
 \addtolength{\belowcaptionskip}{-4.mm}
\resizebox{\textwidth}{!}{
\begin{tabular}{l|l|l|l|l|l|l|l|l|l|l|l|l|l|l|l|l|l|l|l|l|l|l|l|l|l}
\multirow{2}{*}{}          & \multicolumn{1}{c|}{Idle}                                                                                                                                                                      & \multicolumn{6}{c|}{CoNLL03-English}                                                                                                                                   & \multicolumn{6}{c|}{CoNLL02-Spanish}                                                                                                                                    & \multicolumn{6}{c|}{ACE-English}                                                                                                                                       & \multicolumn{6}{c}{ACE-Chinese}                                                                                                                                       \\ \cline{2-26}
          & mins/iter  & 10\%                      & 20\%                      & 30\%                      & 40\%                      & 50\%                      & 100\%                     & 10\%                      & 20\%                      & 30\%                      & 40\%                      & 50\%                      & 100\%                     & 10\%                      & 20\%                      & 30\%                      & 40\%                      & 50\%                      & 100\%                     & 10\%                      & 20\%                      & 30\%                      & 40\%                      & 50\%                      & 100\%                     \\ \hline

\textbf{Full Data}                                                                     & \cellcolor[HTML]{888888}x                                                                                                 & \cellcolor[HTML]{888888}x & \cellcolor[HTML]{888888}x & \cellcolor[HTML]{888888}x & \cellcolor[HTML]{888888}x & \cellcolor[HTML]{888888}x & 92.4                      & \cellcolor[HTML]{888888}x & \cellcolor[HTML]{888888}x & \cellcolor[HTML]{888888}x & \cellcolor[HTML]{888888}x & \cellcolor[HTML]{888888}x & 89.6                      & \cellcolor[HTML]{888888}x & \cellcolor[HTML]{888888}x & \cellcolor[HTML]{888888}x & \cellcolor[HTML]{888888}x & \cellcolor[HTML]{888888}x & 71.9                      & \cellcolor[HTML]{888888}x & \cellcolor[HTML]{888888}x & \cellcolor[HTML]{888888}x & \cellcolor[HTML]{888888}x & \cellcolor[HTML]{888888}x & 69.1                      \\
\textbf{Large}    & 41.6                                                                                                                      & \textbf{90.3}                      & \textbf{92.4}                      & \textbf{93.0}             & \textbf{92.4}                      & 92.4                      & \cellcolor[HTML]{888888}x & 86.9                      & \textbf{88.6}                      & \textbf{89.4}                      & \textbf{89.3}                      & 89.0                      & \cellcolor[HTML]{888888}x & \textbf{67.8}                      & \textbf{71.1}                      & \textbf{70.0}                      & \textbf{72.4}             & \textbf{71.3}                      & \cellcolor[HTML]{888888}x & \textbf{64.8}                      & 67.6                      & \textbf{71.3}                      & 68.7                      & \textbf{71.5}             & \cellcolor[HTML]{888888}x \\
\textbf{FaMIE}     & \textbf{3.4}                                                                                                                      & 90.1                      & 91.7                      & 91.8                      & 91.7                      & \textbf{92.7}                      & \cellcolor[HTML]{888888}x & 86.5                      & 88.2                      & 88.5                      & 88.1                      & \textbf{89.4}                      & \cellcolor[HTML]{888888}x & 67.0                      & 69.3                      & 69.5                      & 68.9                      & 70.6                      & \cellcolor[HTML]{888888}x & 61.3                      & \textbf{67.9}                      & 68.5                      & \textbf{69.8}                      & 69.6                      & \cellcolor[HTML]{888888}x \\
\textbf{FaMIE-A}   & 5.7                                                                                                                      & 89.7                      & 90.8                      & 91.3                      & 91.9                      & 91.7                      & \cellcolor[HTML]{888888}x & \textbf{87.4}                      & 87.2                      & 89.0                      & 87.7                      & 89.1                      & \cellcolor[HTML]{888888}x & 67.2                      & 68.0                      & 69.5                      & 68.9                      & 70.6                      & \cellcolor[HTML]{888888}x & 62.8                      & 66.5                      & 67.9                      & 66.3                      & 69.4                      & \cellcolor[HTML]{888888}x \\
\textbf{FaMIE-AD}  & 5.6                                                                                                                      & 87.0                      & 90.1                      & 90.5                      & 90.7                      & 90.5                      & \cellcolor[HTML]{888888}x & 85.5                      & 86.9                      & 87.7                      & 88.8                      & 88.6                      & \cellcolor[HTML]{888888}x & 64.9                      & 65.4                      & 67.7                      & 66.8                      & 69.1                      & \cellcolor[HTML]{888888}x & 58.1                      & 65.4                      & 66.5                      & 64.8                      & 70.3                      & \cellcolor[HTML]{888888}x \\
\textbf{Random}   & \cellcolor[HTML]{888888}x                                                                                                                       & 86.0                      & 89.1                      & 90.6                      & 91.4                      & 91.9                      & \cellcolor[HTML]{888888}x & 80.8                      & 85.3                      & 88.1                      & 88.7                      & 88.6                      & \cellcolor[HTML]{888888}x & 60.4                      & 64.1                      & 66.9                      & 69.0                      & 67.5                      & \cellcolor[HTML]{888888}x & 48.4                      & 58.2                      & 65.1                      & 65.4                      & 66.6                      & \cellcolor[HTML]{888888}x 
%\textbf{Random}   & 37.0                                                                                                                      & 86.0                      & 89.1                      & 90.6                      & 91.4                      & 91.9                      & \cellcolor[HTML]{888888}x & 80.8                      & 85.3                      & 88.1                      & 88.7                      & 88.6                      & \cellcolor[HTML]{888888}x & 60.4                      & 64.1                      & 66.9                      & 69.0                      & 67.5                      & \cellcolor[HTML]{888888}x & 48.4                      & 58.2                      & 65.1                      & 65.4                      & 66.6                      & \cellcolor[HTML]{888888}x 
\end{tabular}
}
\caption{\small Main model's performance on multilingual NER and ED tasks. ``Idle'' indicate average waiting time of annotators.}
% \textbf{proxy} represents the average time proxy model is retrained and used for data selection per iteration, which is the idle duration of FaMIE-related models. \textbf{target} is retraining time required by target model, taking into account the additional distillation computation (if any).}
\label{tab:1}
\end{table*}

% =================================
% Evaluation
% =================================
\section{Evaluation}

%To provide a full view of different active learning methods for multilingual information extraction problems, we experiment on the following corpora:

%for multilingual sequence labeling problems in IE

\paragraph{Datasets and Hyper-parameters.} To comprehensively evaluate our AL framework FAMIE, we conduct experiments on two IE tasks (i.e., NER and ED) for three languages using four datasets: CoNLL03-English \cite{tjong-kim-sang-de-meulder-2003-introduction} and CoNLL02-Spanish \cite{tjong-kim-sang-2002-introduction} for NER, and ACE-English and ACE-Chinese for ED (i.e., using the multilingual ACE-05 dataset \cite{Walker:05,Nguyen:15b,Nguyen:18a}). The CoNLL datasets cover 4 entity types while 33 event types are annotated in ACE-05 datasets. We follow the standard data splits for train/dev/test portions for each dataset \cite{Li:13,Lai:20event,pouran-ben-veyseh-etal-2021-unleash}.
% detail in appendice ?
% + CoNLL03-English: train/dev/test = 14041/3250/3453 sentences, 4 entity types.
% + CoNLL02-Spanish: train/dev/test = 8323/1915/1517 sentences, 4 entity types
% + ACE-English: train/dev/test = 19240/902/676 sentences, 33 event types
% + ACE-Chinese: train/dev/test = 6841/526/547 sentences, 33 event types

%, which was pre-trained to handle textual data from 100 languages.

%distilled from the pre-trained XLM-R model.

For the main target model $M_{main}$, the full-scale $\textrm{XLM-Roberta}_{large}$ model is used as the encoder. Our framework for AL thus inherits the ability of \textrm{XLM-Roberta} to support more than 100 languages. Also, we employ the compact \textrm{miniLM} architecture (distilled from the pre-trained \textrm{XLM-Roberta}) for the proxy model $M_{prx}$. In all experiments, the main model is trained for 40 epochs while the proxy model is trained for 20 epochs at each iteration. We use the Adam optimizer with batch size of 16 and learning rate of 1$e$-5 to train the models.

%with only 6 layers and 384 dimensions for the hidden vectors

%, gradient clipping of 4.5,

%, making FAMIE the first multilingual AL framework to our knowledge

%and learning rates of 1e-5 and 2e-4 to train the fully fine-tuned and adapter-based models (respectively).

We follow the AL settings in previous work to achieve consistent evaluation \cite{kim-2020-deep,shelmanov-etal-2021-active,liu2020ltp}. Specifically, the unlabeled pool is created by discarding labels from the original training data of each dataset; 2\% of which ($\sim$ 242 sentences) is selected for labeling at each iteration for a total of 25 iterations (examples of the first iteration are randomly sampled to serve as the seed $D_0$). The annotation is simulated by recovering the ground-truth labels of the corresponding instances. The model performance is measured on the test datasets by taking average over 3 runs with different random seeds.

\paragraph{Comparing Data Selection Strategies.} In this experiment, we aim to determine the best data selection strategy for our AL framework. To this end, we perform the standard AL process (i.e., training the full transformer-CRF model with no adapters, selecting data, and annotating data at each iteration) for different data selection strategies to measure performance and time. We focus on English datasets in this experiment. Figure \ref{fig:2} reports the performance across AL iterations of the model for different data selection methods. As can be seen, ``\textit{MNLP}'' is the overall best method for data selection in AL. We will thus leverage MNLP as the data section strategy for the evaluation of FAMIE.

%In particular, with only 12\% and 18\% training data in CoNLL03-English and ACE-English for NER and ED (respectively), ``\textit{MNLP}'' can achieve equivalent results with the model trained on the entire training data.

Also, Figure \ref{fig:2} shows the annotators' idle time (the combined time for model training and data selection) across iterations for each selection strategy. 
The major difference comes from ALPS that has significantly less waiting time than other methods as it does not require model training. However, ALPS's performance is considerably worse than MNLP as a result, especially in early iterations. This demonstrates the importance of training and including the main model during the AL iterations for data section. Importantly, we find that the waiting time of annotators at each iteration is very high in current AL methods (e.g., more than 30 minutes after the first 8 iterations with the MNLP strategy), thus affecting the annotators' productivity.

\paragraph{Performance and Time Efficiency.} To evaluate the performance and time efficiency of FAMIE, Table \ref{tab:1} compares our full proposed framework FAMIE (with proxy model, knowledge distillation, and adapters) with the following baselines:
(i) ``\textbf{Large}'': the best AL baseline from the previous experiment employing the full-scale transformer encoder and MNLP for data selection;
(ii) ``\textbf{Random}'': this is the same as ``\textbf{Large}'', but replaces MNLP with random selection;
(iii) ``\textbf{FAMIE-A}'': this is the proposed framework FAMIE without adapter-based tuning (all parameters from the main model are fine-tuned); and 
(iv) ``\textbf{FAMIE-AD}'': we further remove the knowledge distillation loss from ``\textbf{FAMIE-A}'' in this method. The experiments are done for all four datasets of NER and ED.

The first observation is that FAMIE's performance is only marginally below that of Large despite only using the small proxy network for data selection. Importantly, annotators only have to wait for about 3.4 minutes per AL iteration before they can annotate the next data batch in FAMIE. This is over 10 times faster compared to the standard AL approaches (e.g., in Large). Second, the adapters in FAMIE not only boost the overall performance for AL but also reduce the waiting time for annotators. Also, we note that using adapters, the training time of $M_{main}$ only takes 32 minutes at each iteration (on average). This is reasonable to fit into the time that an annotator needs to spend to label an annotation batch at each AL iteration, thus accommodating our proposal for training the main model during annotation time. Finally, FAMIE-AD performs worst (i.e., similar or even worse than Random) in most cases, which confirms the necessity of our distillation component in FAMIE.

\section{Related Work}
Despite the potential of AL in reducing annotation cost for a target task, most previous AL work focuses on developing data selection strategies to maximize the model performance \cite{wang2014new,sener2017active,ash2019deep,kim-2020-deep,liu2020ltp,margatina2021active}. As such, previous AL methods and frameworks tend to ignore the necessary time to train models and perform data selection at each AL iteration that can be significantly long and hinder annotators' productivity and model performance. To make AL frameworks practical, few recent works have attempted to minimize the model training and data selection time by leveraging simple and non state-of-the-art architectures as the main model, e.g.,  ActiveAnno \cite{wiechmann-etal-2021-activeanno} and Paladin \cite{nghiem-etal-2021-paladin}. However, an issue with these approaches is the inability to exploit recent advances in large-scale language models to achieve optimal performance. In addition, some recent works have also explored large-scale language models for AL \cite{shelmanov-etal-2021-active,yuan-etal-2020-cold}; however, to reduce waiting time for annotators, such methods need to exclude the training of the large models in the AL iterations or employ small models for data selection, thus suffering from a harmful mismatch between the annotated examples and the main models \cite{lowell-etal-2019-practical}.

\section{Conclusion}

We introduce FAMIE, a comprehensive AL framework that supports model creation and data annotation for sequence labeling in multiple languages. FAMIE optimizes the annotators' time by leveraging a small proxy network for data selection and a novel knowledge distillation to synchronize the proxy and main target models for AL. As FAMIE is task-agnostic, we plan to extend FAMIE to cover other NLP tasks in future work.

\section*{Acknowledgement}

This research has been supported by the Army Research Office (ARO) grant W911NF-21-1-0112 and the NSF grant CNS-1747798 to the IUCRC Center for Big Learning. This research is also based upon work supported by the Office of the Director of National Intelligence (ODNI), Intelligence Advanced Research Projects Activity (IARPA), via IARPA Contract No. 2019-19051600006 under the Better Extraction from Text Towards Enhanced Retrieval (BETTER) Program. The views and conclusions contained herein are those of the authors and should not be interpreted as necessarily representing the official policies, either expressed or implied, of ARO, ODNI, IARPA, the Department of Defense, or the U.S. Government. The U.S. Government is authorized to reproduce and distribute reprints for governmental purposes notwithstanding any copyright annotation therein. This document does not contain technology or technical data controlled under either the U.S. International Traffic in Arms Regulations or the U.S. Export Administration Regulations.

% Our experiments suggests that FAMIE can produce sequence labeling models with competitive performance while significantly reducing the waiting time for annotators. 

%(i.e., more than 10 time faster than traditional AL methods)

%FaMIE is a comprehensive active learning toolkit for information extraction tasks in multiple languages. Our framework is optimized for performance of the final model as well as the overall process' efficiency, taking into account both memory and time budgets. 
% We introduce FAMIE, a dual-model AL framework, in which retraining and annotating are paralleled and small proxy model is trained to select appropriate new unlabel data.
%- Our method to achieve competitive performance for NER and ED across multiple language while significantly reduce total time of the whole process.
%- In the future, we intend to ...

\bibliography{anthology,custom}
\bibliographystyle{acl_natbib}

\clearpage

\appendix

\end{document}